%% file: main.tex
\documentclass[10pt,twocolumn,letterpaper]{article}

\usepackage{iccv}      
\usepackage[table]{xcolor}
\usepackage{amssymb}
\usepackage{arydshln}
\usepackage{pifont}
\usepackage{footnote}

\usepackage{multirow}
\usepackage{vcell}

\usepackage{url}

\usepackage{lipsum}

\usepackage[flushleft]{threeparttable}


\definecolor{iccvblue}{rgb}{0.21,0.49,0.74}
\usepackage[pagebackref,breaklinks,colorlinks,allcolors=iccvblue]{hyperref}

\def\paperID{xxxx} 
\def\confName{xxxx}
\def\confYear{xxxx}

\title{WD-FQDet: Multispectral Detection Transformer via Wavelet Decomposition and Frequency-aware Query Learning}

\author{Chunjin Yang, Xiwei Zhang, Yiming Xiao, Fanman Meng*\\
University of Electronic Science and Technology of China,Chengdu, Sichuan 611731, China\\
 \{202321011830,202322011832, ymxiao\}@std.uestc.edu.cn,fmmeng@uestc.edu.cn
\\}

\begin{document}
\maketitle
\input{sec/0_abstract}    
\input{sec/1_intro}

{
    \small
    \bibliographystyle{ieeenat_fullname}
    \bibliography{main}
}

\end{document}

%% file: sec/0_abstract.tex
\begin{abstract}
Infrared-visible object detection improves detection performance by combining complementary features from multispectral images. Existing backbone-specific and backbone-shared approaches still suffer from the problems of severe bias of modality-shared features and the insufficiency of modality-specific features. To address these issues, we propose a novel detection framework WD-FQDet that explicitly decouples modality-shared and modality-specific information from infrared and visible modalities in the new view of low- and high-frequency domains, allowing fusion strategies tailored to their frequency characteristics. Specifically, a low-frequency homogeneity  alignment module is proposed to align modality-shared features across modalities via a cross-modal attention mechanism, and a high-frequency specificity  retention module is proposed to preserve modality-specific features through the multi-scale gradient consistency loss. To reinforce the feature representation in the frequency domain, we propose a hybrid feature enhancement module that incorporates spatial cues. Furthermore, considering that the contributions of homogeneous and modality-specific features to object detection vary across scenarios, we propose a frequency-aware query selection module to dynamically regulate their contributions. Experimental results on the FLIR, LLVIP, and M3FD datasets demonstrate that WD-FQDet achieves state-of-the-art performance across multiple evaluation metrics.

\end{abstract}

%% file: sec/1_intro.tex
\section{Introduction}
\label{sec:intro}

Object detection is a core task in computer vision, playing a pivotal role in critical applications such as autonomous driving \cite{contreras2024survey,zhao2024enhancing,liang2024vehicle}, maritime surveillance \cite{wang2021machine,rekavandi2025guide,cheng2023deep}, and emergency response to natural disasters \cite{geetha2021machine,shafique2022deep,shafique2022deep,chase2023you}. However, under the challenging conditions, such as low illumination, haze and thermal interference,  the performance of conventional object detection methods reduce severely \cite{fu2023lraf}. To this end, multispectral detection frameworks is proposed to fuse infrared (IR) and visible (RGB) information for object detection, drawing significant attention in recent years.
 
\begin{figure}[!t]
\centering
\includegraphics[width=3.3in]{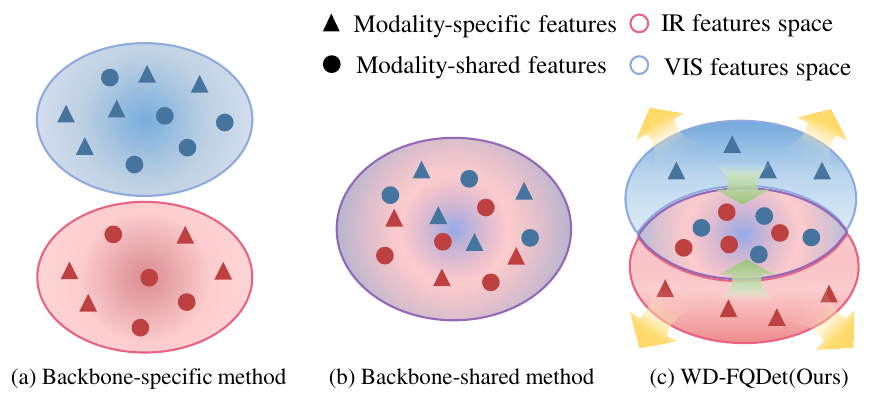}
\caption{(a) The backbone-specific method introduces severe bias in homogeneous information; (b) The backbone-shared method unifies multimodal features, resulting in the loss of modality-specific features; (c) Our approach explicitly decouples homogeneous and modality-specific features, aligning modality-shared features while preserving modality-specific features.}
\label{introduction}
\end{figure}

Currently, there are two main paradigms for infrared-visible object detection: backbone-specific \cite{xing2024ms,dong2024fusion,guo2024dpdetr} and backbone-shared  \cite{zhang2023differential,xiao2024gm} methods. The former adopts  two independent backbone networks to extract infrared and visible features separately, which facilitates capturing modality-specific information but tends to introduce modality homogeneity bias (as shown in Figure \ref{introduction}(a)). In contrast, the latter utilizes a shared backbone to extract features from both modalities, enabling the model to learn cross-modal correlations more effectively. However, mapping multi-modal features into a unified representation space results in the loss of modality-specific information. (as shown in Figure \ref{introduction}(b)). Recent studies in image fusion, such as CDDFuse \cite{zhao2023cddfuse}, indicate that modality-specific features are mainly concentrated in the high-frequency space, while homogeneous information is primarily contained in the low-frequency space. This motivates us to improve infrared-visible object detection by decoupling homogeneous and heterogeneous information in the frequency domain  (as illustrated in Figure \ref{introduction}(c)).

Based on this observation, we propose WD-FQDet, a novel multispectral object detection framework following the backbone-share paradigm. WD-FQDet employs a wavelet transform \cite{farge1992wavelet} to explicitly decompose infrared and visible features into low-frequency and high-frequency components, treating low-frequency components as the modality-shared features and high-frequency components as the modality-specific features. This decomposition enables the decoupling of modality-shared and modality-specific features, allowing us to adopt different fusion strategies tailored to their distinct frequency-domain characteristics. We then introduce a low-frequency homogeneity alignment (LFHA) module that employs cross-modal attention mechanisms to align modality-shared features between infrared and visible images, yielding aligned homogeneous features. Additionally, a high-frequency specificity retention (HFSR) module is proposed to enhance modality-specific features by first extracting edge contours using multi-scale convolutions and the HOG operator, and then applying a multi-scale gradient consistency loss to ensure that the fused modality-specific features fully capture the unique details of both modalities.

Furthermore, we design a hybrid feature enhancement (HFE) module that fuses the refined modality-specific features with visible spatial features to enhance fine details, while also merging the aligned modality-shared features with infrared spatial features to strengthen global structural information. Considering that the contribution of homogeneous and modality-specific features to object detection varies across scenarios (e.g., relying on infrared-specific features in foggy scenes, visible-specific features for small object detection, and modality-shared features for robust detection tasks), we further propose a frequency-aware query selection (FQS) module, enabling the model to dynamically adjust the weight of shared and specific features for optimal cross-modal complementary information utilization.

To validate the effectiveness of WD-FQDet, we conducted experiments on three public datasets—FLIR, LLVIP, and M3FD. The results demonstrate that the proposed method achieves state-of-the-art (SOTA) performance across these datasets, confirming its superiority.

The contributions of this paper are as follows.
\begin{enumerate}
    \item We propose WD-FQDet, a novel multispectral detection transformer based on wavelet decomposition and frequency-aware query learning, enabling efficient fusion of complementary infrared-visible information.
    \item We introduce a wavelet-based feature decomposition and integration module to decouple modality-shared and -specific features in low- and high-frequency domain, including a low-frequency homogeneity alignment module and a high-frequency specificity retention module.
    \item A hybrid feature enhancement module is proposed to integrate homogeneous and modality-specific features with spatial features, enhancing their representation capability. Meanwhile, a frequency-aware query selection module is designed to dynamically adjust the contribution weights of homogeneous and specific features.
    \item Extensive experiments on the FLIR, LLVIP, and M3FD  benchmark datasets show that WD-FQDet achieves state-of-the-art (SOTA) performance, validating its effectiveness and superiority.
\end{enumerate}


\section{related works}
 \begin{figure*}[!t]
\centering
\includegraphics[width=7.1in]{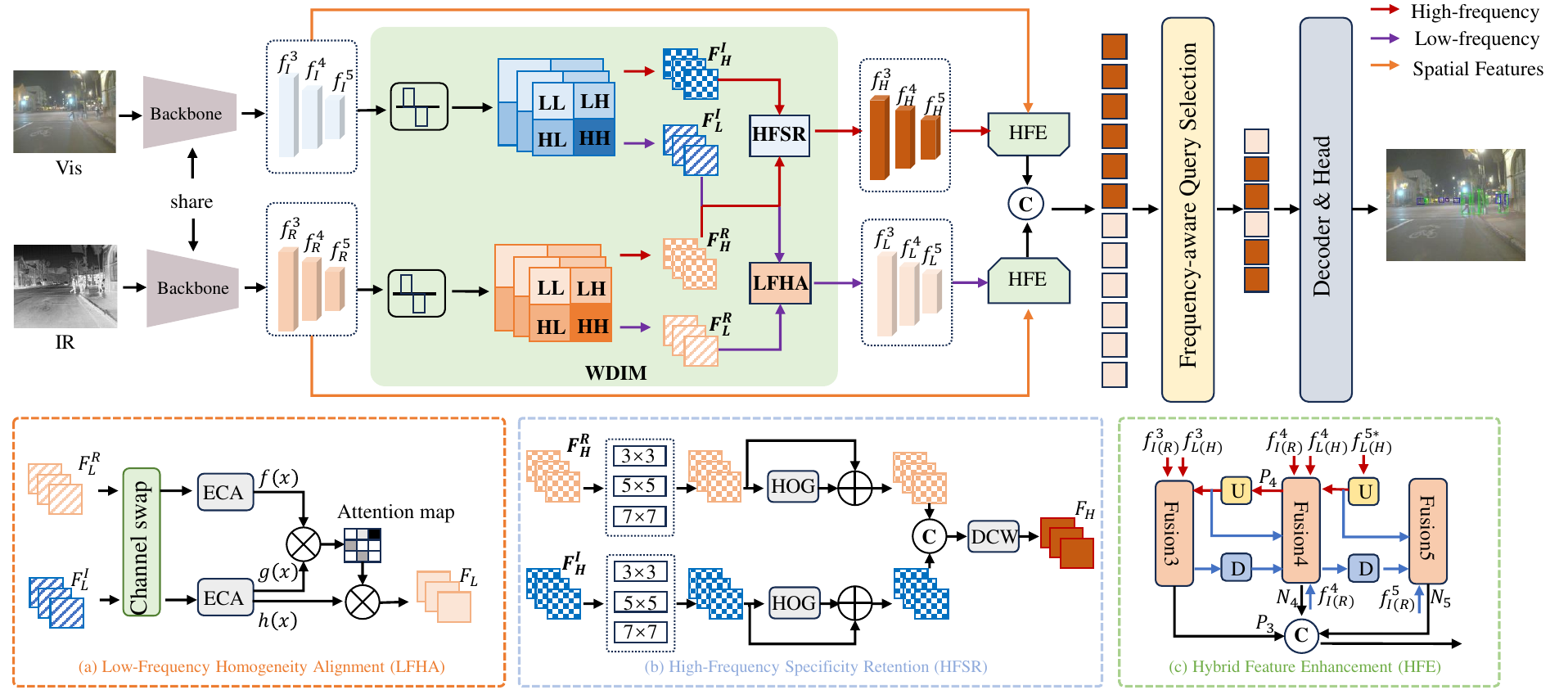}%
\caption{The overall architecture our WD-FQDet. Our WD-FQDet comprises three key modules. First, features extracted by a modality-shared backbone are fed into a wavelet-based feature decomposition and integration module (WDIM), where the LFHA aligns homogeneous information and the HFSR preserves modality-specific features. Next, a hybrid feature enhancement module incorporates spatial features to bolster the representation of frequency-domain features. Finally, the enhanced features are passed into a frequency-aware query selection module, achieving adaptive fusion of homogeneous and modality-specific features.}
\label{overview_pic}
\end{figure*}
\subsection{Infrared-visible Object Detection}
Infrared-visible object detection aims to integrate complementary information from infrared and visible modalities to develop detection performance  under challenging low-light conditions such as poor illumination and occlusion by clouds or fog. Wagner et al. \cite{wagner2016multispectral} first introduced early and late CNN fusion architectures to bolster detection reliability. Deng et al. \cite{deng2021pedestrian} proposed a multilayer fusion network that generates multiscale feature maps from both visible and infrared channels, thereby improving detection accuracy in low-light environments. Konig et al. \cite{konig2017fully} presented a fully convolutional fusion RPN network that concatenates infrared and visible features, demonstrating that mid-level fusion can yield superior results.

Moreover, several works have incorporated illumination as prior knowledge by employing illumination-aware networks to adaptively fuse visible and thermal features \cite{li2019illumination,liu2021deep}. Guan et al. \cite{guan2019fusion} further proposed illumination-aware modules that enable object detectors to adjust fusion weights based on predicted illumination conditions. In addition to these approaches, TFDet \cite{zhang2024tfdet} introduces a segmentation head as a constraint and puts forward a novel target-aware fusion strategy for multispectral pedestrian detection, while Junjie Guo et al. \cite{guo2024dpdetr} decouple positional and class information by introducing a multispectral decoder within an end-to-end transformer-based framework. Despite the significant achievements of existing methods, they either suffer from an excessive bias toward cross-modal homogeneous information or incur the loss of modality-specific features. Therefore, we explicitly decouple homogeneous and modality-specific  features, and employ distinct fusion enhancement strategies tailored to their inherent frequency domain characteristics.
\subsection{Wavelet Decomposition in Deep Learning}
Wavelet Decomposition \cite{farge1992wavelet}, as a powerful tool for signal analysis and processing, has seen widespread adoption in neural network architectures.

In the field of image denoising, Hwang et al. \cite{hwang2024wavedh} employed the discrete wavelet transform  to decompose embedded low- and high-frequency components and computed attention based on the low-frequency parts, significantly mitigating noise effects. Tian et al. \cite{tian2023multi} designed dynamic convolution blocks that adaptively adjust parameters and, in combination with a wavelet-based enhancement block to isolate noisy frequency bands, reconstructed clean images via residual blocks. 

In the field of image fusion, Bhavana et al. \cite{bhavana2015multi} utilized DWT to fuse brain regions with varying activity levels, reducing color distortion, while Zhang et al. \cite{zhang2024enhanced} introduced wavelet transform to independently train high- and low-frequency components, thereby enhancing fusion accuracy and quality using only two remote sensing images. 

In the domain of image segmentation, Zhou et al. \cite{zhou2023xnet} proposed a wavelet-based high- and low-frequency fusion model, XNet, which supports both fully supervised and semi-supervised semantic segmentation, effectively improving biomedical segmentation performance. Yang et al. \cite{yang2024sffnet} presented a spatial and frequency domain fusion network that fully leverages both spatial and frequency domain information, achieving significant results in remote sensing image segmentation. 

Moreover, Shahaf et al. \cite{finder2024wavelet} introduced a novel network layer based on the wavelet transform, named WTConv, which can serve as a drop-in replacement in existing architectures. However, the potential of the wavelet transform in infrared-visible object detection remains largely unexplored. Given its remarkable success in modeling high- and low-frequency features, we likewise employ the wavelet transform for the feature decomposition of both infrared and visible modalities.

\section{Method}
\subsection{Formulation of infrared-visible object detection}
The objective of infrared-visible object detection is to enhance detection performance by effectively leveraging the complementary information from both modalities, while simultaneously using the modality-shared features to suppress noise. Assuming that the features extracted from infrared and visible images adhere to a Gaussian distribution, the shared semantic characteristics—such as object contours and spatial positions—in the same scene should exhibit highly consistent statistical distributions. Let the probability distribution of homogeneous features in the latent space be defined as:
\begin{equation}
    \begin{cases}p_{shared}^{IR}\sim\mathcal{N}(\mu_{sh}^{IR},\Sigma_{sh}^{IR})\\p_{shared}^{RGB}\sim\mathcal{N}(\mu_{sh}^{RGB},\Sigma_{sh}^{RGB})&\end{cases}.
\end{equation}
To align homogeneous features, we constrain the similarity between their distributions using Kullback–Leibler (KL) divergence and align their mean features using the L2 norm. The model's optimization objective is given by:
\begin{equation}
    \begin{aligned}
        \min\ &\mathbb{E}_{x\sim p_{shared}} \Big[ \|\mu_{sh}^{IR} - \mu_{sh}^{RGB}\|_{2}^{2} \Big] \\
        &+ \lambda_{1} D_{KL}(p_{shared}^{IR} \parallel p_{shared}^{RGB}).
    \end{aligned}
\end{equation}
For modality-specific features, a distribution independence constraint is imposed:
\begin{equation}
    \begin{cases}p_{specific}^{IR}\sim\mathcal{N}(\mu_{sp}^{IR},\Sigma_{sp}^{IR})\\p_{specific}^{RGB}\sim\mathcal{N}(\mu_{sp}^{RGB},\Sigma_{sp}^{RGB})&\end{cases}.
\end{equation}
To ensure that the fused  features fully retain the unique characteristics of both modalities, we aim to enforce the orthogonality of their covariance matrix through the following constraint: 
\begin{equation}
    \mathrm{Tr}(\Sigma_{sp}^{IR}\Sigma_{sp}^{{RGB}^T})=0\quad\text{and}\quad\max\|\mu_{sp}^{IR}-\mu_{sp}^{RGB}\|_2^2,
\end{equation}
where $\mu_{space}^{mod}$ represents the mean vector, $\Sigma_{space}^{mod}$ represents the covariance matrix,  where $space\in \{shared,specific\}$, $mod\in \{IR,RGB\}$.
\begin{figure}[!t]
\centering
\includegraphics[width=3in]{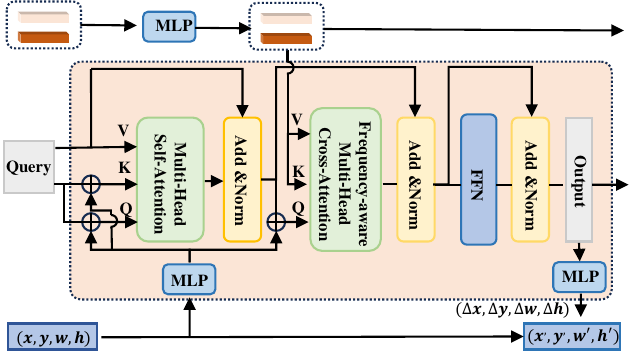}
\caption{Illustration of the frequency-aware query-selection module. The high-frequency and low-frequency features are linearly mapped and their weights are adaptively adjusted before being concatenated. These features are then computed with the query, updated by the self-attention mechanism, through the Frequency-aware Multi-Head Cross-Attention process.}
\label{head}
\end{figure}
\subsection{Overview}
Figure \ref{overview_pic} illustrates the overall architecture of WD-FQDet, which consists of three main modules: a wavelet-based feature decomposition and integration module, a hybrid feature enhancement module, and a frequency-aware query selection module. For a given pair of infrared and visible images, a shared backbone first extracts multi-scale spatial features. These features are then decomposed into low-frequency homogeneous and high-frequency heterogeneous  components via the wavelet transform. The low-frequency homogeneity alignment module synchronizes the shared features, while the high-frequency specificity retention module reserves modality-specific features from both modalities. Subsequently, the hybrid enhancement module reinforces the frequency-domain representations. Finally, the enhanced features are fed into the frequency-aware query selection module, which adaptively balances the contributions of modality-share and -specific features, updating the queries that are processed by the RT-DETR \cite{zhao2024detrs} detection head to generate bounding boxes and classification scores.

\subsection{Wavelet-based Feature Decomposition and Integration Module}

The wavelet-based feature decomposition module first decomposes the spatial features of infrared and visible images into four components using haar wavelet \cite{farge1992wavelet} transform: $LL$, $LH$, $HL$, and $HH$. 

Following this process, the three high-frequency components are concatenated and passed through a point-wise convolution to reduce the dimensionality, resulting in the final high-frequency and low-frequency features. After applying the wavelet decomposition to the spatial domain features of both infrared and visible images, we obtain the high-frequency infrared features $F^R_H = \{f^{R_3}_H, f^{R_4}_H, f^{R_5}_H\}$ and visible features $F^I_H = \{f^{I_3}_H, f^{I_4}_H, f^{I_5}_H\}$, as well as the low-frequency infrared features $F^R_L = \{f^{R_3}_L, f^{R_4}_L, f^{R_5}_L\}$ and visible features $F^I_L = \{f^{I_3}_L, f^{I_4}_L, f^{I_5}_L\}$, for further details, please refer to the Appendix.

\subsubsection{Low-Frequency Homogeneity Alignment }
To align the homogeneous information between the infrared and visible modalities, we designed a low-frequency homogeneity alignment (LFHA) module, as illustrated in Figure \ref{overview_pic}(a). Specifically, the low-frequency homogeneous features of both modality are first fused through channel swapping for preliminary integration,
then enhanced via an efficient channel attention mechanism \cite{wang2020eca} to reinforce salient features. Finally, a cross-modal attention mechanism is applied to capture the inter-modal correlations between infrared and visible features, thereby achieving precise alignment of homogeneous information:
\begin{equation}
    f_L^i = \operatorname{Softmax} \left( \frac{ f_i \big( f_L^{R_i} \big) g_i \big( f_L^{I_i} \big)^T }{\sqrt{D}} \right) h_i \big( f_L^{I_i} \big),
\end{equation}
where $f$, $g$, and $h$ denote linear layers, $D$ represents the feature encoding dimension, and $i\in \{3,4,5\}$.
\subsubsection{High-Frequency Specificity Retention}
To preserve the modality-specific features of infrared and visible images, we designed a high-frequency specificity retention (HFSR) module, as illustrated in Figure \ref{overview_pic}(b). Initially, multi-scale convolutions are applied to the high-frequency modality-specific features of each modality to capture detail information at various granularities,
subsequently, structured edge features are extracted using the histogram of oriented gradients (HOG) \cite{dalal2005histograms} and fused with the original high-frequency features via residual connections:
\begin{equation}
f_H^{R_i\ast}=\textbf{HOG}(f_H^{R_i})+f_H^{R_i},f_H^{I_i\ast}=\textbf{HOG}(f_H^{I_i})+f_H^{I_i}.
\end{equation}
Subsequently, the enhanced features $f_H^{R_i\ast}$ and $f_H^{I_i\ast}$  are concatenated and processed via depthwise separable convolution \cite{chollet2017xception} for channel compression, yielding the final fused features $f_H^i$. To ensure that the fused high-frequency features adequately preserve the modality-specific characteristics of both infrared and visible images, we introduce a multi-scale gradient consistency loss:

\begin{equation}
\mathcal{L}_{\text{grad}} = \frac{1}{N} \sum_{k=1}^3 \Bigl[ \lVert f_H^i - \mathbf{G_k}(f_H^{R_i}) \rVert_1 + \lVert f_H^i - \mathbf{G_k}(f_H^{I_i}) \rVert_1 \Bigr],
\label{Lgrad}
\end{equation}
where $\mathbf{G_k}$ is a multi-scale gradient feature extraction function, implemented using the Sobel operator \cite{kanopoulos1988design} and discrete dilated convolution \cite{yu2015multi}.
Through the aforementioned feature fusion and loss constraints, the HFSR module effectively integrates the modality-specific features of infrared and visible images, while simultaneously enhancing the representation of fine details such as edges and contours.

\subsection{Hybrid Feature Enhancement Module}
We propose a hybrid feature enhancement (HFE) module that reinforces  structural and  detail representations by integrating aligned low-frequency homogeneous features with infrared spatial cues and merging fused high-frequency modality-specific features with visible spatial cues, as illustrated in Figure \ref{overview_pic}(c).


For brevity, we denote  $f_{L(H)}^i$ to representing  either aligned low-frequency homogeneous features $f_L^i$ or fused high-frequency modality-specific features  $f_H^i$, and $f_{I(R)}^i$ to representing  either visible spatial features $f_I^i$ or infrared spatial features $f_R^i$. A self-attention mechanism \cite{vaswani2017attention} is then applied $f_{L(H)}^{5}$ to get $f_{L(H)}^{5\ast}$.

We further design a PANet-inspired feature fusion module comprising two branches: a top-down FPN and a bottom-up PAN. First, top-down upsampling of \( f_{L(H)}^{5\ast} \) produces high-resolution semantic features. Then, bottom-up downsampling of \( f_{L(H)}^4 \) is aggregated with \( f_{I(R)}^4 \) to construct a feature pyramid that integrates frequency and spatial cues. The FPN process is defined as:
\begin{equation}
\begin{aligned}
 P_4 &= \mathbf{Fusion4}\Big(\mathbf{U}\big(f_{L(H)}^{5\ast}\big),\, f_{L(H)}^4,\, f_{I(R)}^4\Big), \\
 P_3 &= \mathbf{Fusion3}\Big(\mathbf{U}(P_4),\, f_{L(H)}^3,\, f_{I(R)}^3\Big),
\end{aligned}
\end{equation}
where \( \mathbf{U}(\cdot) \) denotes a \( 1 \times 1 \) convolution for upsampling, and \( \mathbf{Fusion3} \) and \( \mathbf{Fusion4} \) are fusion blocks comprising \( N \) RepBlocks with element-wise addition.

The PAN process is formulated as:
\begin{equation}
\begin{aligned}
 N_4 &= \mathbf{Fusion4}\Big(\mathbf{D}(P_3),\, (P_4)_{1 \times 1},\, f_{I(R)}^4\Big), \\
 N_5 &= \mathbf{Fusion5}\Big(\mathbf{D}(N_4),\, (f_{L(H)}^{5\ast})_{1 \times 1},\, f_{I(R)}^5\Big),
\end{aligned}
\end{equation}
where \( \mathbf{D}(\cdot) \) is a \( 3 \times 3 \) convolution with stride 2, and \( (\cdot)_{1 \times 1} \) represents the output of a \( 1 \times 1 \) convolution.

The module ultimately yields enhanced low-frequency homogeneous features \( \{P_3^L, N_4^L, N_5^L\} \) and high-frequency heterogeneous features \( \{P_3^H, N_4^H, N_5^H\} \).

\begin{table*}[]
\centering
\setlength{\tabcolsep}{6pt}
\renewcommand{\arraystretch}{1}
\caption{The AP performance comparison on the FLIR dataset includes both single-modal and multimodal methods, with the best results highlighted in \textbf{Bold}, the second-best \underline{underlined}, and those marked with $\dagger$ indicating that we re-trained the experiments on the dataset using the official code, for experimental details, please refer to the appendix.}
\begin{tabular}{cccccccccc}
\toprule[1pt]
Method  & Venue & Backbone & Modal &  People $\uparrow$ & Car $\uparrow$ & Bicycle $\uparrow$ & $mAP_{50}$$\uparrow$ & $mAP_{75}$ $\uparrow$ & mAP$\uparrow$  \\ \hline
\multicolumn{1}{c|}{RT-DETR$^{\dagger}$ \cite{zhao2024detrs}}  & CVPR'24  & \multicolumn{1}{c|}{ResNet50}     & \multicolumn{1}{c|}{IR}     & 83.8   & 90.2 & \multicolumn{1}{c|}{66.6} & 80.2  & 39.2  & 43.2 \\ \cline{4-4}
\multicolumn{1}{c|}{RT-DETR$^{\dagger}$ \cite{zhao2024detrs}}  & CVPR'24  & \multicolumn{1}{c|}{ResNet50}     & \multicolumn{1}{c|}{RGB}    & 69.2   & 82.1 & \multicolumn{1}{c|}{53.6} & 68.3  & 26.0    & 33.0   \\ \hline
\multicolumn{1}{c|}{SDNet \cite{zhang2021sdnet}}    & IJCV'21  & \multicolumn{1}{c|}{-}             & \multicolumn{1}{c|}{}       & 81.0   & 87.3 & \multicolumn{1}{c|}{64.2} & 77.5  & 33.1  &-    \\
\multicolumn{1}{c|}{CDDFuse \cite{zhao2023cddfuse}}  & CVPR'23  & \multicolumn{1}{c|}{Restormer}             & \multicolumn{1}{c|}{}       & 82.3   & 87.2 & \multicolumn{1}{c|}{72.9} & 80.8  & 39.4  & -    \\
\multicolumn{1}{c|}{MetaF \cite{zhao2023metafusion}}    & CVPR'23  & \multicolumn{1}{c|}{-}             & \multicolumn{1}{c|}{}       & 83.3   & 89.2 & \multicolumn{1}{c|}{71.1} & 81.4  & 40.7  & -    \\
\multicolumn{1}{c|}{LRAF-Net \cite{fu2023lraf}} & TNNLS'23 & \multicolumn{1}{c|}{CSP53} & \multicolumn{1}{c|}{}       & 83.3   & 88.8 & \multicolumn{1}{c|}{69.7} & 80.6  & 41.0  & 42.8 \\
\multicolumn{1}{c|}{SegMiF \cite{liu2023multi}}   & ICCV'23  & \multicolumn{1}{c|}{ViT}             & \multicolumn{1}{c|}{}       & 85.3   & 86.9 & \multicolumn{1}{c|}{72.8} & 81.5  & 40.9  & -    \\
\multicolumn{1}{c|}{TarDAL \cite{liu2022target}}  & CVPR'22  & \multicolumn{1}{c|}{CSPD53} & \multicolumn{1}{c|}{}       & 85.1   & 85.3 & \multicolumn{1}{c|}{69.3} & 79.9  & 37.9  & -    \\
\multicolumn{1}{c|}{DDFM \cite{zhao2023ddfm}}     & ICCV'23  & \multicolumn{1}{c|}{-}             & \multicolumn{1}{c|}{IR+RGB} & 84.5   & 87.9 & \multicolumn{1}{c|}{71.5} & 81.2  & 40.2  & -    \\
\multicolumn{1}{c|}{CSSA \cite{cao2023multimodal}}     & CVPR'23  & \multicolumn{1}{c|}{ResNet50}             & \multicolumn{1}{c|}{}       & 83.2   & 86.7 & \multicolumn{1}{c|}{68.6} & 79.4  & 37.2  & 41.3 \\
\multicolumn{1}{c|}{TFDet \cite{zhang2024tfdet}}    & TNNLS'24 & \multicolumn{1}{c|}{CSP53} & \multicolumn{1}{c|}{}       & 85.2   & 87.5 & \multicolumn{1}{c|}{71.9} & 81.7  & 41.3  & \underline{46.6} \\
\multicolumn{1}{c|}{GM-DETR$^{\dagger}$ \cite{xiao2024gm}}  & CVPR'24  & \multicolumn{1}{c|}{ResNet50}     & \multicolumn{1}{c|}{}       & \underline{87.9}   & \underline{91.3} & \multicolumn{1}{c|}{66.6} & 81.9  & \underline{42.8}  & 45.4 \\
\multicolumn{1}{c|}{FD2Net \cite{li2024fd2}}   & AAAI'25  & \multicolumn{1}{c|}{ResNet50}     & \multicolumn{1}{c|}{}       & 85.3   & 89.9 & \multicolumn{1}{c|}{\underline{73.2}} & \underline{82.9}  & 42.5  & -     \\
\rowcolor{blue!8} \multicolumn{1}{c|}{Ours}     & -         & \multicolumn{1}{c|}{ResNet50}     & \multicolumn{1}{c|}{}       & \textbf{89.3}   & \textbf{92.4} & \multicolumn{1}{c|}{\textbf{79.3}} & \textbf{87.0}  & \textbf{50.1}  & \textbf{50.2} \\ \hline
\end{tabular}
\label{FLIR_exp}
\end{table*}

\begin{figure}[!t]
\centering
\includegraphics[width=3.3in]{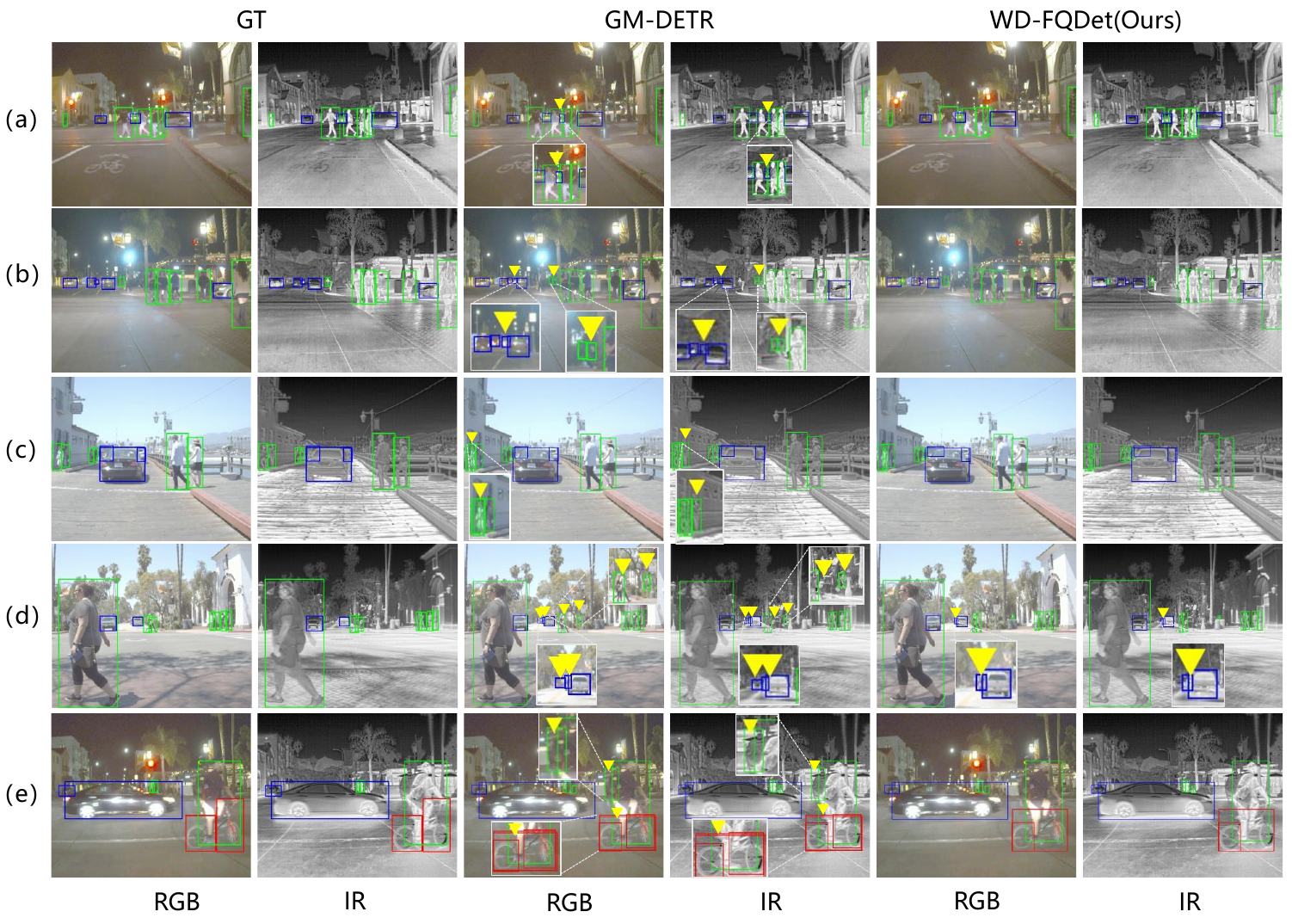}%
\caption{Visualization results on the FLIR dataset. False positives and missed detections are marked with \textcolor{yellow}{$\blacktriangledown$}. Compared to GM-DETR \cite{xiao2024gm}, our method demonstrates significant performance improvement, particularly for small and weak targets in complex scenes.}
\label{visible_FLIR}
\end{figure}
\subsection{Frequency-Aware Query Selection Module}

Considering that the contributions of homogeneous and modality-specific features to object detection vary across scenarios, we propose a frequency-aware query selection module to effectively fuse homogeneous and modality-specific information. This module dynamically balances the contributions of each feature based on the test scenario, thereby facilitating the integration of homogeneous and modality-specific features.

As illustrated in Figure ~\ref{head}, the module first maps enhanced low-frequency homogeneous features ($\mathbf{F_L}$) and high-frequency modality-specific features ($\mathbf{F_H}$) through an MLP to obtain weighting coefficients, which are then used to scale the corresponding features:
\begin{equation} 
\begin{aligned}   
\mathbf{F_H} &\leftarrow \mathbf{F_H} \times \text{MLP}(\text{Flatten}(\mathbf{F_H})), \\    
\mathbf{F_L} &\leftarrow \mathbf{F_L} \times \text{MLP}(\text{Flatten}(\mathbf{F_L})).
\end{aligned}
\end{equation}
The weighted, flattened features are concatenated, and the top $K$ highest-scoring features are selected as the initial queries:
\begin{equation}
    z = \text{TopK}\Big(\text{Linear}\Big(\text{Cat}\big(\text{Flatten}(\mathbf{F_H}),\,\text{Flatten}(\mathbf{F_L})\big)\Big)\Big).
\end{equation}

Subsequently, a multi-head self-attention module is applied. Let $A_q = (x_q, y_q, w_q, h_q)$ denote the $q$-th anchor, with its content and positional queries represented by $z_q$ and $P_q \in \mathbb{R}^D$, respectively (where $D$ is the decoder embedding dimension). The positional query is generated as:
\begin{equation}
\mathbf{P_q} = \text{MLP}(\mathbf{A_q}).
\end{equation}
The self-attention computation is then expressed as:
\begin{equation} 
\begin{aligned}   
  \mathbf{Q_q} &= z_q + \mathbf{P_q}, \quad \mathbf{K_q} = z_q + \mathbf{P_q}, \quad \mathbf{V_q} = z_q, \\
  \hat{z_q}& = \text{Softmax}\Big(\frac{\mathbf{Q_q}\mathbf{K_q}^\top}{\sqrt{D}}\Big) \mathbf{V_q}.
\end{aligned}
\end{equation}

We denote high-frequency modality-specific features as $\{h\}_i^I$, low-frequency modality-share features as $\{l\}_i^I$, and use the normalized center $\hat{b_q}$ of $b_q$ as the 2D reference point for consistent positional comparison. Inspired by DETR \cite{carion2020end}, we define the frequency-aware cross-attention(FACA) module as:
\begin{equation}
\begin{aligned}FACA \left(\boldsymbol{\hat{z}}_{q}, \hat{\boldsymbol{p}}_{q}, \left\{\boldsymbol{l^i}\right\}_{i=1}^{I}, \left\{\boldsymbol{h^i}\right\}_{i=1}^{I} \right) =  \sum_{m=1}^{M} \boldsymbol{W}_{m} \\\Bigg[ \sum_{i=1}^{I} \sum_{k=1}^{K} A_{m i q k} \cdot \boldsymbol{W}_{m}^{\prime} \Big( \boldsymbol{l}^{i} \left( \phi\left( \hat{\boldsymbol{p}}_{q} \right) + \Delta \boldsymbol{p}_{m i q k} \right)  \Big) &+ \\ \sum_{i=1}^{I} \sum_{k=1}^{K} A_{m i q k} \cdot \boldsymbol{W}_{m}^{\prime} \Big( \boldsymbol{h}^{i} \left( \phi\left( \hat{\boldsymbol{p}}_{q} \right) + \Delta \boldsymbol{p}_{m i q k} \right) \Big) \Bigg].\end{aligned}
\end{equation}
where $m$, $i$, and $k$ index the attention head, feature level, and sampling point, respectively; $\Delta p_{miqk}$ and $A_{miqk}$ denote the sampling offset and normalized attention weight for the $k$-th sampling point at level $i$ in head $m$.

After several iterations of the frequency-aware selection, the queries can robustly extract features tailored to the test scene. These optimized queries are then passed to the RT-DETR \cite{zhao2024detrs} detection head for precise object localization and classification. In this manner, the model dynamically adjusts its focus on frequency-domain features, enhancing detection accuracy and robustness in complex or variable environments.

\subsection{Traing Loss}
During training, in addition to the original detection task loss, we incorporate our multi-scale gradient consistency loss (as shown in Equation \ref{Lgrad}) to ensure that the proposed HFSR fully extracts the modality-specific features of infrared and visible images. The overall loss function is defined as follows:
\begin{equation}
\mathcal{L}_{\text{total}} = \mathcal{L}_{\text{box}}+\mathcal{L}_{\text{cls}}+\mathcal{L}_{\text{grad}},
\end{equation}
where $\mathcal{L}_{\text{box}}$ and $\mathcal{L}_{\text{cls}}$ employ the same loss functions as those used in RT-DETR \cite{zhao2024detrs}.
\section{Experiment}
\subsection{Datasets and Metric}
We conducted experiments on three datasets covering diverse scenarios: FLIR \cite{FLIR}, LLVIP \cite{jia2021llvip}, and M3FD \cite{liu2022target}. Standard COCO AP \cite{lin2014microsoft} was used as the evaluation metric.
\subsubsection{FLIR} FLIR dataset is a challenging multispectral object detection dataset that includes both daytime and nighttime scenes. We use the aligned version with a resolution of 640×512 pixels. This dataset contains 5,142 well-aligned visible-infrared image pairs, of which 4,129 pairs are used for training and 1,013 pairs are used for testing. It covers three object categories: ``person'', ``car" and ``bicycle".
\subsubsection{LLVIP} The LLVIP dataset is a multispectral object detection dataset captured under low-light conditions.  It includes 12,025 aligned infrared-visible image pairs as the training set and 3,463 aligned pairs as the testing set. The image resolution is 1,024×1,280. Most of the images are taken under dim lighting conditions and contain only one object category: ``pedestrian". 

\subsubsection{M3FD} The M3FD dataset  serves as a benchmark for multispectral object detection. It comprises six object categories: ``person", ``car", ``bus", ``motorcycle", ``light", and ``truck".  Since the dataset does not provide an official split, we divided it based on different scenes into a training set with 3,368 image pairs and a validation set with 831 image pairs following \cite{zhang2024tfdet}.

\subsection{Experiment Details}
We utilized ResNet-50 \cite{he2016deep} as the modality-shared backbone and extracted multi-scale features from its 3-rd, 4-th, and 5-th layers. Haar wavelet transformation was employed for wavelet decomposition. The encoder in the multi-scale frequency domain feature enhancement module consists of 1 layer, with 8 attention heads. The RT-DETR \cite{zhao2024detrs} decoder head includes 6 layers of Transformers, utilizing 300 queries, with a feature embedding dimension of 256. We set the input image size to 640 × 640 for training and testing on FLIR, LLVIP, and M3FD datasets

\begin{table}[]
\setlength{\tabcolsep}{1pt}
\renewcommand{\arraystretch}{1}
\caption{The AP performance comparison on the LLVIP dataset includes both single-modal and multimodal methods, with the best results highlighted in \textbf{Bold} and the second-best \underline{underlined}.}
\begin{tabular}{ccccc}
\toprule[1pt]
Model       & Backbone     & Modal & $mAP_{50}$$\uparrow$ & mAP$\uparrow$  \\ \hline 
DDQ-DETR \cite{zhang2023dense}    & \multicolumn{1}{c|}{ResNet50}     & \multicolumn{1}{c|}{\multirow{2}{*}{IR}}     & 87.0    & 45.1 \\
DINO \cite{zhang2022dino}         & \multicolumn{1}{c|}{ResNet50}     & \multicolumn{1}{c|}{}                        & 96.6  & 62.9 \\ \cline{3-3}
DDQ-DETR \cite{zhang2023dense}    & \multicolumn{1}{c|}{ResNet50}     & \multicolumn{1}{c|}{\multirow{2}{*}{RGB}}    & 86.1  & 46.7 \\
DINO \cite{zhang2022dino}        & \multicolumn{1}{c|}{ResNet50}     & \multicolumn{1}{c|}{}                        & 91.6  & 53.8 \\ \hline
ProbEn \cite{chen2022multimodal}       & \multicolumn{1}{c|}{ResNet50}     & \multicolumn{1}{c|}{\multirow{9}{*}{IR+RGB}} & 93.4  & 51.5 \\
CSSA \cite{cao2023multimodal}         & \multicolumn{1}{c|}{ResNet50}     & \multicolumn{1}{c|}{}                        & 94.3  & 59.2 \\
RSDet \cite{zhao2024removal}        & \multicolumn{1}{c|}{ResNet50}     & \multicolumn{1}{c|}{}                        & 95.8  & 61.3 \\
FD2Net \cite{li2024fd2}        & \multicolumn{1}{c|}{ResNet50}     & \multicolumn{1}{c|}{}                        & 96.2  & - \\
LRAF-Net \cite{fu2023lraf}     & \multicolumn{1}{c|}{CSP53} & \multicolumn{1}{c|}{}                        & \underline{97.9}  & \underline{66.3} \\
MS-DETR \cite{xing2024ms}      & \multicolumn{1}{c|}{ResNet50}     & \multicolumn{1}{c|}{}                        & \underline{97.9}  & 66.1 \\
CFT \cite{qingyun2021cross}           & \multicolumn{1}{c|}{CSP53} & \multicolumn{1}{c|}{}                        & 97.5  & 63.6 \\
Fusion-Mamba \cite{dong2024fusion} & \multicolumn{1}{c|}{CSP53} & \multicolumn{1}{c|}{}                        & 96.8  & 62.8 \\
\rowcolor{blue!8}Ours         & \multicolumn{1}{c|}{ResNet50}     & \multicolumn{1}{c|}{}                        & \textbf{98.2}  & \textbf{66.9} \\ \bottomrule[1pt]
\end{tabular}
\label{LLVIP_ana}
\end{table}


\subsection{Comparison on the FLIR Dataset}
We conducted experiments on the FLIR dataset and compared the results with existing state-of-the-art methods. As shown in Table \ref{FLIR_exp}, compared to the best single-modality method, RT-DETR \cite{zhao2024detrs}, our approach achieved improvements of \textbf{6.8\%} in $mAP_{50}$, \textbf{10.9\%} in $mAP_{75}$, and \textbf{7\%} in $mAP$. When compared to multimodal methods, our performance also improved: $mAP_{50}$ increased by \textbf{4.1\%} compared to FD2Net \cite{li2024fd2}, $mAP_{75}$ increased by \textbf{7.3\%} compared to GM-DETR \cite{xiao2024gm}, and $mAP$ improved by \textbf{3.6\%} compared to TFDet \cite{zhang2024tfdet}. Sample detection results are illustrated in Figure \ref{visible_FLIR}, demonstrating that our method significantly outperforms existing methods, especially for small and weak targets in complex backgrounds.

\subsection{Comparison on the LLVIP Dataset}
Experiments on the LLVIP pedestrian detection dataset demonstrate that our method significantly outperforms existing models. As shown in Table \ref{LLVIP_ana}, compared to the leading single-modal DINO \cite{zhangdino}, our approach achieves an \textbf{4\%} $mAP$ gain, and a \textbf{1.6\%} boost in $mAP_{50}$. Against LRAF-Net \cite{fu2023lraf}, the best multi-modal model, our method improves $mAP$ by \textbf{0.6\%}, and $mAP_{50}$ by \textbf{0.3\%}. Moreover, compared with MS-DETR \cite{xing2024ms}, which employs a similar DETR \cite{carion2020end} detection head and ResNet50 backbone, our method attains a \textbf{0.8\%} $mAP$ enhancement. These results underscore the efficacy and superiority of our approach in single-category object detection. 


\begin{table}[t]
\centering
\setlength{\tabcolsep}{1pt}
\renewcommand{\arraystretch}{1}
\caption{The AP performance comparison on the M3FD dataset includes both single-modal and multimodal methods, with the best results highlighted in \textbf{Bold} and the second-best \underline{underlined}. }
\begin{tabular}{ccccc}
\toprule[1pt]
Model     & Backbone     & Modal& $mAP_{50}$$\uparrow$ & mAP$\uparrow$  \\ \hline
Yolov5 \cite{yolov5}    & \multicolumn{1}{c|}{CSP53}    & \multicolumn{1}{c|}{RGB}                     & 60.2  & 36.1 \\ \cline{3-3}
Yolov5 \cite{yolov5}    & \multicolumn{1}{c|}{CSP53}    & \multicolumn{1}{c|}{IR}                      & 57.2  & 34.9 \\ \hline
DIVFusion \cite{tang2023divfusion} & \multicolumn{1}{c|}{-}        & \multicolumn{1}{c|}{\multirow{7}{*}{IR+RGB}} & 60.8  & 37.1 \\
PSFusion \cite{tang2023rethinking}  & \multicolumn{1}{c|}{ResNet50} & \multicolumn{1}{c|}{}                        & 61.1  & 38   \\
AUIF \cite{zhao2021efficient}      & \multicolumn{1}{c|}{-}        & \multicolumn{1}{c|}{}                        & 62    & 38.3 \\
CDDFuse \cite{zhao2023cddfuse}   & \multicolumn{1}{c|}{Restormer}        & \multicolumn{1}{c|}{}                        & 61.9  & 38.6 \\
TarDAL \cite{liu2022target}    & \multicolumn{1}{c|}{CSP53}    & \multicolumn{1}{c|}{}                        & 61.9  & 39.1 \\
TFDet \cite{zhang2024tfdet}     & \multicolumn{1}{c|}{CSP53}    & \multicolumn{1}{c|}{}                        & \underline{64.8}  & \underline{41.0} \\
\rowcolor{blue!8} Ours      & \multicolumn{1}{c|}{ResNet50} & \multicolumn{1}{c|}{}                        & \textbf{73.7}  & \textbf{46.4} \\ \bottomrule[1pt]
\end{tabular}
\label{M3FD}
\end{table}

\subsection{Comparison on the M3FD Dataset}
We conducted experiments on the autonomous driving dataset MFD and compared our approach with existing state-of-the-art methods, as shown in Table \ref{M3FD}. Compared to YOLO, the top-performing single-modal model, our approach achieves a \textbf{13.5\%} improvement in $mAP_{50}$ and a \textbf{10.3\%} gain in $mAP$. Similarly, relative to best multi-modal method TFDet \cite{zhang2024tfdet}, our method yields an \textbf{8.9\%} boost in $mAP_{50}$ and a \textbf{5.4\%} increase in $mAP$. Moreover, when compared with PSFusion \cite{tang2023rethinking}, which also employs ResNet50 as the backbone, our method attains a \textbf{12.6\%} enhancement in $mAP_{50}$ and a \textbf{6.4\%} improvement in $mAP$.

\subsection{Computational Cost Comparison}



We evaluated the computational efficiency of the proposed method on the FLIR dataset as shown in Table \ref{conputing}. Our method achieves state-of-the-art performance with moderate computational requirements. Compared to previous works like CrossFormer \cite{lee2024crossformer}  and CFT \cite{qingyun2021cross}, our model significantly reduces parameter count while maintaining competitive computational efficiency. Though requiring more computation than lightweight models like LRAF-Net \cite{fu2023lraf}, the increased cost brings substantial performance gains.
 
\begin{table}[t]
\centering
\caption{A comparison between our WD-FQDet and existing approaches in terms of model parameters and computational complexity.}
\setlength{\tabcolsep}{6pt}
\renewcommand{\arraystretch}{1}
\begin{tabular}{cccc}
\toprule[1pt]
Methods     & mAP $\uparrow$   & GFLOPs$\downarrow$ & Params$\downarrow$ \\ \hline 
TSFAdet \cite{yuan2022translation}      & 73.1 & 109.8  & 104.7M \\
CFT \cite{qingyun2021cross}          & 77.5  & 154.7  & 206.0M   \\
C2Former \cite{yuan2024c}     & 74.2  & 177.7  & 120.8M \\
CrossFormer \cite{lee2024crossformer}  & 79.3  & 361.7 & 340.0M   \\
LRAF-Net \cite{fu2023lraf}     & 80.5  & \textbf{40.5}   & \textbf{18.8M}  \\
\rowcolor{blue!8}Ours         & \textbf{87.0}  & 162.9 & 60.7M \\ 
\bottomrule[1pt] 
\label{conputing}
\end{tabular}
\end{table}

\begin{figure}[ht]
\centering
\includegraphics[width=3.3in]{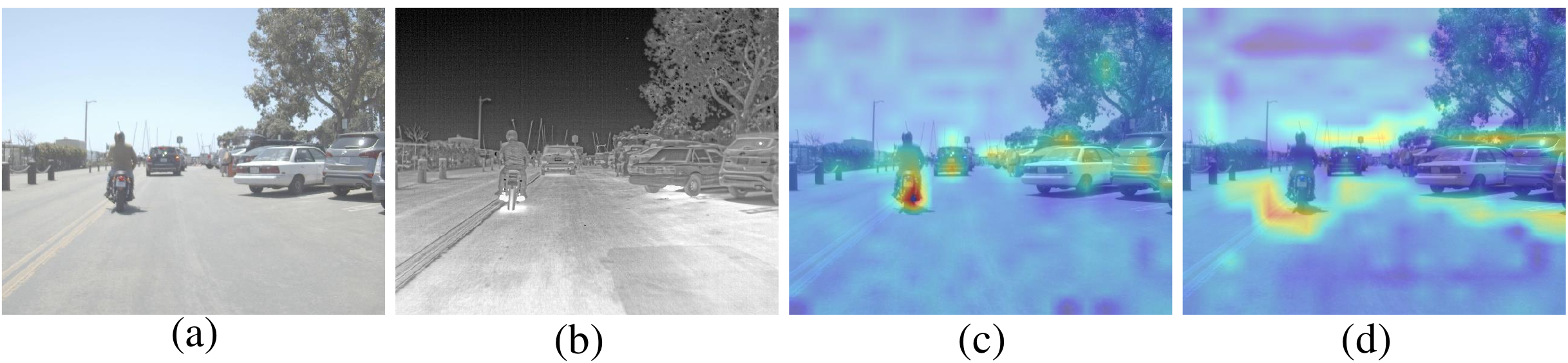}
\caption{(a) represents the visible image; (b) represents the corresponding infrared image; (c) represents the heatmap of low-frequency homogeneous features; and (d) represents the heatmap of high-frequency specific features.}
\label{hotmap}
\end{figure}

\subsection{Ablation Study}

\paragraph{\textbf{Wavelet-based Feature  Integration Module}}


We replaced the high-frequency  specificity retention (HFSR) and low-frequency homogeneity alignment (LFHA) modules with a design that concatenates features along the channel dimension, followed by a 1×1 convolution for channel reduction. As shown in Table \ref{ablation}, the experimental results indicate that removing either module significantly degrades performance, thereby validating the efficacy of our proposed method.
Additionally, we visualized the heatmaps of the fused low-frequency homogeneous features and high-frequency specific features, as shown in Figure \ref{hotmap}(c) and \ref{hotmap}(d). It can be observed that the low-frequency homogeneous features focus more on overall structural information, while the high-frequency specific features emphasize details such as edge contours. This confirms the effectiveness of our decoupling in the wavelet domain.


\paragraph{\textbf{Hybrid Feature Enhancement Module}}

We validated the effectiveness of the hybrid feature enhancement (HFE) module through an ablation study in which we removed the spatial-domain features, forcing the model to rely solely on frequency-domain features. As shown in Table \ref{ablation}, on the FLIR dataset, this modification resulted in a 2.3\% $mAP$ drop, underscoring the critical role of the HFE.

\paragraph{\textbf{Frequency-Aware Query Selection Module}}

To validate the effectiveness of the frequency-aware query selection (FQS) Module, we conducted an experiment in which the high-frequency specific and low-frequency specific features enhanced by the HFE were directly summed, adopting a fixed fusion strategy. As shown in Table \ref{ablation}, the results reveal a 3\% drop in $mAP$, confirming the efficacy of the FQS module.

\begin{table}[t]
\centering
\caption{Ablation experiments on the key modules of our proposed method.}
\setlength{\tabcolsep}{3pt} 

\begin{tabular}{cccc|ccc}
\toprule[1pt]
\multicolumn{4}{c|}{Module} & \multicolumn{3}{c}{Metric} \\ \hline
HFSR  & LFHA  & HFE  & FQS  & $mAP_{50}$ $\uparrow$   & $mAP_{75}$ $\uparrow$   & mAP $\uparrow$    \\ \hline
$\checkmark$     & $\checkmark$     & $\checkmark$    & $\ding{55}$      & 85.1    & 44.9    & 46.8   \\
$\checkmark$     & $\checkmark$     & $\ding{55}$     & $\checkmark$    & 85.8    & 46.6    & 47.9   \\
$\checkmark$     & $\ding{55}$      & $\checkmark$    & $\checkmark$    & 82.5    & 42.6    & 43.2   \\
$\ding{55}$     & $\checkmark$     & $\checkmark$    & $\checkmark$    & 83.8    & 42.9    & 44.5   \\
\rowcolor{blue!8} $\checkmark$     & $\checkmark$     & $\checkmark$    & $\checkmark$    & \textbf{87.0}      & \textbf{50.1}    & \textbf{50.2}   \\ 
\bottomrule[1pt]

\end{tabular}
\label{ablation}
\end{table}


\section{Conclusion}
In this paper, we propose a novel approach that leverages wavelet transforms to decompose infrared and visible features into modality-shared and modality-specific components in low-frequency and high-frequency domain. A low-frequency homogeneity alignment module synchronizes the shared information, while a high-frequency specificity retention module preserves the unique details of each modality, effectively decoupling the two feature types. Moreover, a hybrid feature enhancement module integrates spatial features to bolster the structural information and enhance the fine details. Finally, acknowledging that the reliance on homogeneous versus modality-specific features varies across scenarios, we design a frequency-aware query selection module to achieve adaptive fusion. Our method demonstrates significant performance gains on the FLIR, LLVIP, and M3FD datasets.




%% file: main.bbl
\begin{thebibliography}{63}
\providecommand{\natexlab}[1]{#1}
\providecommand{\url}[1]{\texttt{#1}}
\expandafter\ifx\csname urlstyle\endcsname\relax
  \providecommand{\doi}[1]{doi: #1}\else
  \providecommand{\doi}{doi: \begingroup \urlstyle{rm}\Url}\fi

\bibitem[Bhavana and Krishnappa(2015)]{bhavana2015multi}
V Bhavana and HK Krishnappa.
\newblock Multi-modality medical image fusion using discrete wavelet transform.
\newblock \emph{Procedia Computer Science}, 70:\penalty0 625--631, 2015.

\bibitem[Cao et~al.(2023)Cao, Bin, Hamari, Blasch, and Liu]{cao2023multimodal}
Yue Cao, Junchi Bin, Jozsef Hamari, Erik Blasch, and Zheng Liu.
\newblock Multimodal object detection by channel switching and spatial attention.
\newblock In \emph{Proceedings of the IEEE/CVF Conference on Computer Vision and Pattern Recognition}, pages 403--411, 2023.

\bibitem[Carion et~al.(2020)Carion, Massa, Synnaeve, Usunier, Kirillov, and Zagoruyko]{carion2020end}
Nicolas Carion, Francisco Massa, Gabriel Synnaeve, Nicolas Usunier, Alexander Kirillov, and Sergey Zagoruyko.
\newblock End-to-end object detection with transformers.
\newblock In \emph{European conference on computer vision}, pages 213--229. Springer, 2020.

\bibitem[Chase~Jr et~al.(2023)Chase~Jr, Gnam, Crassidis, and Dantu]{chase2023you}
Timothy Chase~Jr, Chris Gnam, John Crassidis, and Karthik Dantu.
\newblock You only crash once: Improved object detection for real-time, sim-to-real hazardous terrain detection and classification for autonomous planetary landings.
\newblock \emph{arXiv preprint arXiv:2303.04891}, 2023.

\bibitem[Chen et~al.(2022)Chen, Shi, Ye, Mertz, Ramanan, and Kong]{chen2022multimodal}
Yi-Ting Chen, Jinghao Shi, Zelin Ye, Christoph Mertz, Deva Ramanan, and Shu Kong.
\newblock Multimodal object detection via probabilistic ensembling.
\newblock In \emph{European Conference on Computer Vision}, pages 139--158. Springer, 2022.

\bibitem[Cheng et~al.(2023)Cheng, Zhu, and Wu]{cheng2023deep}
Shuxiao Cheng, Yishuang Zhu, and Shaohua Wu.
\newblock Deep learning based efficient ship detection from drone-captured images for maritime surveillance.
\newblock \emph{Ocean engineering}, 285:\penalty0 115440, 2023.

\bibitem[Chollet(2017)]{chollet2017xception}
Fran{\c{c}}ois Chollet.
\newblock Xception: Deep learning with depthwise separable convolutions.
\newblock In \emph{Proceedings of the IEEE conference on computer vision and pattern recognition}, pages 1251--1258, 2017.

\bibitem[Contreras et~al.(2024)Contreras, Jain, Bhatt, Banerjee, and Hashemi]{contreras2024survey}
Marcelo Contreras, Aayush Jain, Neel~P Bhatt, Arunava Banerjee, and Ehsan Hashemi.
\newblock A survey on 3d object detection in real time for autonomous driving.
\newblock \emph{Frontiers in Robotics and AI}, 11:\penalty0 1212070, 2024.

\bibitem[Dalal and Triggs(2005)]{dalal2005histograms}
Navneet Dalal and Bill Triggs.
\newblock Histograms of oriented gradients for human detection.
\newblock In \emph{2005 IEEE computer society conference on computer vision and pattern recognition (CVPR'05)}, pages 886--893. Ieee, 2005.

\bibitem[Deng et~al.(2021)Deng, Tian, Huang, Xiong, and Bi]{deng2021pedestrian}
Qing Deng, Wei Tian, Yuyao Huang, Lu Xiong, and Xin Bi.
\newblock Pedestrian detection by fusion of rgb and infrared images in low-light environment.
\newblock In \emph{2021 IEEE 24th International Conference on Information Fusion (FUSION)}, pages 1--8. IEEE, 2021.

\bibitem[Dong et~al.(2024)Dong, Zhu, Lin, Luo, Shen, Liu, Zhang, Guo, and Zhang]{dong2024fusion}
Wenhao Dong, Haodong Zhu, Shaohui Lin, Xiaoyan Luo, Yunhang Shen, Xuhui Liu, Juan Zhang, Guodong Guo, and Baochang Zhang.
\newblock Fusion-mamba for cross-modality object detection.
\newblock \emph{arXiv preprint arXiv:2404.09146}, 2024.

\bibitem[Farge et~al.(1992)]{farge1992wavelet}
Marie Farge et~al.
\newblock Wavelet transforms and their applications to turbulence.
\newblock \emph{Annual review of fluid mechanics}, 24\penalty0 (1):\penalty0 395--458, 1992.

\bibitem[Finder et~al.(2024)Finder, Amoyal, Treister, and Freifeld]{finder2024wavelet}
Shahaf~E Finder, Roy Amoyal, Eran Treister, and Oren Freifeld.
\newblock Wavelet convolutions for large receptive fields.
\newblock In \emph{European Conference on Computer Vision}, pages 363--380. Springer, 2024.

\bibitem[FLIR()]{FLIR}
FLIR.
\newblock Flir thermal dataset for algorithm training.
\newblock \url{https://www.flir.in/oem/adas/adas-dataset-form/}.
\newblock 2018.

\bibitem[Fu et~al.(2023)Fu, Wang, Duan, Xiao, Dian, Li, and Li]{fu2023lraf}
Haolong Fu, Shixun Wang, Puhong Duan, Changyan Xiao, Renwei Dian, Shutao Li, and Zhiyong Li.
\newblock Lraf-net: Long-range attention fusion network for visible--infrared object detection.
\newblock \emph{IEEE Transactions on Neural Networks and Learning Systems}, 2023.

\bibitem[Geetha et~al.(2021)Geetha, Abhishek, and Akshayanat]{geetha2021machine}
S Geetha, CS Abhishek, and CS Akshayanat.
\newblock Machine vision based fire detection techniques: A survey.
\newblock \emph{Fire technology}, 57\penalty0 (2):\penalty0 591--623, 2021.

\bibitem[Guan et~al.(2019)Guan, Cao, Yang, Cao, and Yang]{guan2019fusion}
Dayan Guan, Yanpeng Cao, Jiangxin Yang, Yanlong Cao, and Michael~Ying Yang.
\newblock Fusion of multispectral data through illumination-aware deep neural networks for pedestrian detection.
\newblock \emph{Information Fusion}, 50:\penalty0 148--157, 2019.

\bibitem[Guo et~al.(2024)Guo, Gao, Liu, and Meng]{guo2024dpdetr}
Junjie Guo, Chenqiang Gao, Fangcen Liu, and Deyu Meng.
\newblock Dpdetr: Decoupled position detection transformer for infrared-visible object detection.
\newblock \emph{arXiv preprint arXiv:2408.06123}, 2024.

\bibitem[He et~al.(2016)He, Zhang, Ren, and Sun]{he2016deep}
Kaiming He, Xiangyu Zhang, Shaoqing Ren, and Jian Sun.
\newblock Deep residual learning for image recognition.
\newblock In \emph{Proceedings of the IEEE conference on computer vision and pattern recognition}, pages 770--778, 2016.

\bibitem[Hwang et~al.(2024)Hwang, Han, Jung, and Jeon]{hwang2024wavedh}
Seongmin Hwang, Daeyoung Han, Cheolkon Jung, and Moongu Jeon.
\newblock Wavedh: Wavelet sub-bands guided convnet for efficient image dehazing.
\newblock \emph{arXiv preprint arXiv:2404.01604}, 2024.

\bibitem[Jia et~al.(2021)Jia, Zhu, Li, Tang, and Zhou]{jia2021llvip}
Xinyu Jia, Chuang Zhu, Minzhen Li, Wenqi Tang, and Wenli Zhou.
\newblock Llvip: A visible-infrared paired dataset for low-light vision.
\newblock In \emph{Proceedings of the IEEE/CVF international conference on computer vision}, pages 3496--3504, 2021.

\bibitem[Jocher()]{yolov5}
G. Jocher.
\newblock Yolov5 by ultralytics.
\newblock \url{https://github.com/ultralytics/yolov5}.
\newblock 2020.

\bibitem[Kanopoulos et~al.(1988)Kanopoulos, Vasanthavada, and Baker]{kanopoulos1988design}
Nick Kanopoulos, Nagesh Vasanthavada, and Robert~L Baker.
\newblock Design of an image edge detection filter using the sobel operator.
\newblock \emph{IEEE Journal of solid-state circuits}, 23\penalty0 (2):\penalty0 358--367, 1988.

\bibitem[Konig et~al.(2017)Konig, Adam, Jarvers, Layher, Neumann, and Teutsch]{konig2017fully}
Daniel Konig, Michael Adam, Christian Jarvers, Georg Layher, Heiko Neumann, and Michael Teutsch.
\newblock Fully convolutional region proposal networks for multispectral person detection.
\newblock In \emph{Proceedings of the IEEE conference on computer vision and pattern recognition workshops}, pages 49--56, 2017.

\bibitem[Lee et~al.(2024)Lee, Park, and Park]{lee2024crossformer}
Seungik Lee, Jaehyeong Park, and Jinsun Park.
\newblock Crossformer: Cross-guided attention for multi-modal object detection.
\newblock \emph{Pattern Recognition Letters}, 179:\penalty0 144--150, 2024.

\bibitem[Li et~al.(2019)Li, Song, Tong, and Tang]{li2019illumination}
Chengyang Li, Dan Song, Ruofeng Tong, and Min Tang.
\newblock Illumination-aware faster r-cnn for robust multispectral pedestrian detection.
\newblock \emph{Pattern Recognition}, 85:\penalty0 161--171, 2019.

\bibitem[Li et~al.(2024)Li, Wang, Hu, Li, Ni, Zhao, and Wang]{li2024fd2}
Ke Li, Di Wang, Zhangyuan Hu, Shaofeng Li, Weiping Ni, Lin Zhao, and Quan Wang.
\newblock Fd2-net: Frequency-driven feature decomposition network for infrared-visible object detection.
\newblock \emph{arXiv preprint arXiv:2412.09258}, 2024.

\bibitem[Liang et~al.(2024)Liang, Ma, Zhao, Xie, Hua, Zhang, and Zhang]{liang2024vehicle}
Liang Liang, Haihua Ma, Le Zhao, Xiaopeng Xie, Chengxin Hua, Miao Zhang, and Yonghui Zhang.
\newblock Vehicle detection algorithms for autonomous driving: A review.
\newblock \emph{Sensors}, 24\penalty0 (10):\penalty0 3088, 2024.

\bibitem[Lin et~al.(2014)Lin, Maire, Belongie, Hays, Perona, Ramanan, Doll{\'a}r, and Zitnick]{lin2014microsoft}
Tsung-Yi Lin, Michael Maire, Serge Belongie, James Hays, Pietro Perona, Deva Ramanan, Piotr Doll{\'a}r, and C~Lawrence Zitnick.
\newblock Microsoft coco: Common objects in context.
\newblock In \emph{Computer vision--ECCV 2014: 13th European conference, zurich, Switzerland, September 6-12, 2014, proceedings, part v 13}, pages 740--755. Springer, 2014.

\bibitem[Liu et~al.(2022)Liu, Fan, Huang, Wu, Liu, Zhong, and Luo]{liu2022target}
Jinyuan Liu, Xin Fan, Zhanbo Huang, Guanyao Wu, Risheng Liu, Wei Zhong, and Zhongxuan Luo.
\newblock Target-aware dual adversarial learning and a multi-scenario multi-modality benchmark to fuse infrared and visible for object detection.
\newblock In \emph{Proceedings of the IEEE/CVF conference on computer vision and pattern recognition}, pages 5802--5811, 2022.

\bibitem[Liu et~al.(2023)Liu, Liu, Wu, Ma, Liu, Zhong, Luo, and Fan]{liu2023multi}
Jinyuan Liu, Zhu Liu, Guanyao Wu, Long Ma, Risheng Liu, Wei Zhong, Zhongxuan Luo, and Xin Fan.
\newblock Multi-interactive feature learning and a full-time multi-modality benchmark for image fusion and segmentation.
\newblock In \emph{Proceedings of the IEEE/CVF international conference on computer vision}, pages 8115--8124, 2023.

\bibitem[Liu et~al.(2021)Liu, Lam, Zhao, and Qiu]{liu2021deep}
Tianshan Liu, Kin-Man Lam, Rui Zhao, and Guoping Qiu.
\newblock Deep cross-modal representation learning and distillation for illumination-invariant pedestrian detection.
\newblock \emph{IEEE Transactions on Circuits and Systems for Video Technology}, 32\penalty0 (1):\penalty0 315--329, 2021.

\bibitem[Qingyun et~al.(2021)Qingyun, Dapeng, and Zhaokui]{qingyun2021cross}
Fang Qingyun, Han Dapeng, and Wang Zhaokui.
\newblock Cross-modality fusion transformer for multispectral object detection.
\newblock \emph{arXiv preprint arXiv:2111.00273}, 2021.

\bibitem[Rekavandi et~al.(2025)Rekavandi, Xu, Boussaid, Seghouane, Hoefs, and Bennamoun]{rekavandi2025guide}
Aref~Miri Rekavandi, Lian Xu, Farid Boussaid, Abd-Krim Seghouane, Stephen Hoefs, and Mohammed Bennamoun.
\newblock A guide to image-and video-based small object detection using deep learning: Case study of maritime surveillance.
\newblock \emph{IEEE Transactions on Intelligent Transportation Systems}, 2025.

\bibitem[Shafique et~al.(2022)Shafique, Cao, Khan, Asad, and Aslam]{shafique2022deep}
Ayesha Shafique, Guo Cao, Zia Khan, Muhammad Asad, and Muhammad Aslam.
\newblock Deep learning-based change detection in remote sensing images: A review.
\newblock \emph{Remote Sensing}, 14\penalty0 (4):\penalty0 871, 2022.

\bibitem[Tang et~al.(2023{\natexlab{a}})Tang, Xiang, Zhang, Gong, and Ma]{tang2023divfusion}
Linfeng Tang, Xinyu Xiang, Hao Zhang, Meiqi Gong, and Jiayi Ma.
\newblock Divfusion: Darkness-free infrared and visible image fusion.
\newblock \emph{Information Fusion}, 91:\penalty0 477--493, 2023{\natexlab{a}}.

\bibitem[Tang et~al.(2023{\natexlab{b}})Tang, Zhang, Xu, and Ma]{tang2023rethinking}
Linfeng Tang, Hao Zhang, Han Xu, and Jiayi Ma.
\newblock Rethinking the necessity of image fusion in high-level vision tasks: A practical infrared and visible image fusion network based on progressive semantic injection and scene fidelity.
\newblock \emph{Information Fusion}, 99:\penalty0 101870, 2023{\natexlab{b}}.

\bibitem[Tian et~al.(2023)Tian, Zheng, Zuo, Zhang, Zhang, and Zhang]{tian2023multi}
Chunwei Tian, Menghua Zheng, Wangmeng Zuo, Bob Zhang, Yanning Zhang, and David Zhang.
\newblock Multi-stage image denoising with the wavelet transform.
\newblock \emph{Pattern Recognition}, 134:\penalty0 109050, 2023.

\bibitem[Vaswani et~al.(2017)Vaswani, Shazeer, Parmar, Uszkoreit, Jones, Gomez, Kaiser, and Polosukhin]{vaswani2017attention}
Ashish Vaswani, Noam Shazeer, Niki Parmar, Jakob Uszkoreit, Llion Jones, Aidan~N Gomez, {\L}ukasz Kaiser, and Illia Polosukhin.
\newblock Attention is all you need.
\newblock \emph{Advances in neural information processing systems}, 30, 2017.

\bibitem[Wagner et~al.(2016)Wagner, Fischer, Herman, Behnke, et~al.]{wagner2016multispectral}
J{\"o}rg Wagner, Volker Fischer, Michael Herman, Sven Behnke, et~al.
\newblock Multispectral pedestrian detection using deep fusion convolutional neural networks.
\newblock In \emph{ESANN}, pages 509--514, 2016.

\bibitem[Wang et~al.(2020)Wang, Wu, Zhu, Li, Zuo, and Hu]{wang2020eca}
Qilong Wang, Banggu Wu, Pengfei Zhu, Peihua Li, Wangmeng Zuo, and Qinghua Hu.
\newblock Eca-net: Efficient channel attention for deep convolutional neural networks.
\newblock In \emph{Proceedings of the IEEE/CVF conference on computer vision and pattern recognition}, pages 11534--11542, 2020.

\bibitem[Wang et~al.(2021)Wang, Rajesh, Mercilin~Raajini, Kritika, Kavinkumar, and Shah]{wang2021machine}
Yu Wang, G Rajesh, X Mercilin~Raajini, N Kritika, A Kavinkumar, and Syed Bilal~Hussain Shah.
\newblock Machine learning-based ship detection and tracking using satellite images for maritime surveillance.
\newblock \emph{Journal of Ambient Intelligence and Smart Environments}, 13\penalty0 (5):\penalty0 361--371, 2021.

\bibitem[Xiao et~al.(2024)Xiao, Meng, Wu, Xu, He, and Li]{xiao2024gm}
Yiming Xiao, Fanman Meng, Qingbo Wu, Linfeng Xu, Mingzhou He, and Hongliang Li.
\newblock Gm-detr: Generalized muiltispectral detection transformer with efficient fusion encoder for visible-infrared detection.
\newblock In \emph{Proceedings of the IEEE/CVF Conference on Computer Vision and Pattern Recognition}, pages 5541--5549, 2024.

\bibitem[Xing et~al.(2024)Xing, Yang, Wang, Zhang, Liang, Zhang, and Zhang]{xing2024ms}
Yinghui Xing, Shuo Yang, Song Wang, Shizhou Zhang, Guoqiang Liang, Xiuwei Zhang, and Yanning Zhang.
\newblock Ms-detr: Multispectral pedestrian detection transformer with loosely coupled fusion and modality-balanced optimization.
\newblock \emph{IEEE Transactions on Intelligent Transportation Systems}, 2024.

\bibitem[Yang et~al.(2024)Yang, Yuan, and Li]{yang2024sffnet}
Yunsong Yang, Genji Yuan, and Jinjiang Li.
\newblock Sffnet: A wavelet-based spatial and frequency domain fusion network for remote sensing segmentation.
\newblock \emph{IEEE Transactions on Geoscience and Remote Sensing}, 2024.

\bibitem[Yu and Koltun(2015)]{yu2015multi}
Fisher Yu and Vladlen Koltun.
\newblock Multi-scale context aggregation by dilated convolutions.
\newblock \emph{arXiv preprint arXiv:1511.07122}, 2015.

\bibitem[Yuan and Wei(2024)]{yuan2024c}
Maoxun Yuan and Xingxing Wei.
\newblock C 2 former: Calibrated and complementary transformer for rgb-infrared object detection.
\newblock \emph{IEEE Transactions on Geoscience and Remote Sensing}, 2024.

\bibitem[Yuan et~al.(2022)Yuan, Wang, and Wei]{yuan2022translation}
Maoxun Yuan, Yinyan Wang, and Xingxing Wei.
\newblock Translation, scale and rotation: cross-modal alignment meets rgb-infrared vehicle detection.
\newblock In \emph{European Conference on Computer Vision}, pages 509--525. Springer, 2022.

\bibitem[Zhang and Ma(2021)]{zhang2021sdnet}
Hao Zhang and Jiayi Ma.
\newblock Sdnet: A versatile squeeze-and-decomposition network for real-time image fusion.
\newblock \emph{International Journal of Computer Vision}, 129\penalty0 (10):\penalty0 2761--2785, 2021.

\bibitem[Zhang et~al.()Zhang, Li, Liu, Zhang, Su, Zhu, Ni, and Shum]{zhangdino}
Hao Zhang, Feng Li, Shilong Liu, Lei Zhang, Hang Su, Jun Zhu, Lionel Ni, and Heung-Yeung Shum.
\newblock Dino: Detr with improved denoising anchor boxes for end-to-end object detection.
\newblock In \emph{The Eleventh International Conference on Learning Representations}.

\bibitem[Zhang et~al.(2022)Zhang, Li, Liu, Zhang, Su, Zhu, Ni, and Shum]{zhang2022dino}
Hao Zhang, Feng Li, Shilong Liu, Lei Zhang, Hang Su, Jun Zhu, Lionel~M Ni, and Heung-Yeung Shum.
\newblock Dino: Detr with improved denoising anchor boxes for end-to-end object detection.
\newblock \emph{arXiv preprint arXiv:2203.03605}, 2022.

\bibitem[Zhang et~al.(2023{\natexlab{a}})Zhang, Li, Zhang, Zhang, Xu, Zhang, and Wang]{zhang2023differential}
Ruiheng Zhang, Lu Li, Qi Zhang, Jin Zhang, Lixin Xu, Baomin Zhang, and Binglu Wang.
\newblock Differential feature awareness network within antagonistic learning for infrared-visible object detection.
\newblock \emph{IEEE Transactions on Circuits and Systems for Video Technology}, 2023{\natexlab{a}}.

\bibitem[Zhang et~al.(2023{\natexlab{b}})Zhang, Wang, Wang, Pang, Lyu, Zhang, Luo, and Chen]{zhang2023dense}
Shilong Zhang, Xinjiang Wang, Jiaqi Wang, Jiangmiao Pang, Chengqi Lyu, Wenwei Zhang, Ping Luo, and Kai Chen.
\newblock Dense distinct query for end-to-end object detection.
\newblock In \emph{Proceedings of the IEEE/CVF conference on computer vision and pattern recognition}, pages 7329--7338, 2023{\natexlab{b}}.

\bibitem[Zhang et~al.(2024{\natexlab{a}})Zhang, Li, Tan, and Li]{zhang2024enhanced}
Xingjian Zhang, Shuang Li, Zhenyu Tan, and Xinghua Li.
\newblock Enhanced wavelet based spatiotemporal fusion networks using cross-paired remote sensing images.
\newblock \emph{ISPRS Journal of Photogrammetry and Remote Sensing}, 211:\penalty0 281--297, 2024{\natexlab{a}}.

\bibitem[Zhang et~al.(2024{\natexlab{b}})Zhang, Zhang, Wang, Ying, Sheng, Yu, Li, and Shen]{zhang2024tfdet}
Xue Zhang, Xiaohan Zhang, Jiangtao Wang, Jiacheng Ying, Zehua Sheng, Heng Yu, Chunguang Li, and Hui-Liang Shen.
\newblock Tfdet: Target-aware fusion for rgb-t pedestrian detection.
\newblock \emph{IEEE Transactions on Neural Networks and Learning Systems}, 2024{\natexlab{b}}.

\bibitem[Zhao et~al.(2024{\natexlab{a}})Zhao, Tang, Shen, Supeni, and Rahim]{zhao2024enhancing}
Ruixin Zhao, Sai~Hong Tang, Jiazheng Shen, Eris Elianddy~Bin Supeni, and Sharafiz~Abdul Rahim.
\newblock Enhancing autonomous driving safety: a robust traffic sign detection and recognition model tsd-yolo.
\newblock \emph{Signal Processing}, 225:\penalty0 109619, 2024{\natexlab{a}}.

\bibitem[Zhao et~al.(2024{\natexlab{b}})Zhao, Yuan, Jiang, Wang, and Wei]{zhao2024removal}
Tianyi Zhao, Maoxun Yuan, Feng Jiang, Nan Wang, and Xingxing Wei.
\newblock Removal and selection: Improving rgb-infrared object detection via coarse-to-fine fusion.
\newblock \emph{arXiv preprint arXiv:2401.10731}, 2024{\natexlab{b}}.

\bibitem[Zhao et~al.(2023{\natexlab{a}})Zhao, Xie, Zhao, He, and Lu]{zhao2023metafusion}
Wenda Zhao, Shigeng Xie, Fan Zhao, You He, and Huchuan Lu.
\newblock Metafusion: Infrared and visible image fusion via meta-feature embedding from object detection.
\newblock In \emph{Proceedings of the IEEE/CVF Conference on Computer Vision and Pattern Recognition}, pages 13955--13965, 2023{\natexlab{a}}.

\bibitem[Zhao et~al.(2024{\natexlab{c}})Zhao, Lv, Xu, Wei, Wang, Dang, Liu, and Chen]{zhao2024detrs}
Yian Zhao, Wenyu Lv, Shangliang Xu, Jinman Wei, Guanzhong Wang, Qingqing Dang, Yi Liu, and Jie Chen.
\newblock Detrs beat yolos on real-time object detection.
\newblock In \emph{Proceedings of the IEEE/CVF conference on computer vision and pattern recognition}, pages 16965--16974, 2024{\natexlab{c}}.

\bibitem[Zhao et~al.(2021)Zhao, Xu, Zhang, Liang, Zhang, and Liu]{zhao2021efficient}
Zixiang Zhao, Shuang Xu, Jiangshe Zhang, Chengyang Liang, Chunxia Zhang, and Junmin Liu.
\newblock Efficient and model-based infrared and visible image fusion via algorithm unrolling.
\newblock \emph{IEEE Transactions on Circuits and Systems for Video Technology}, 32\penalty0 (3):\penalty0 1186--1196, 2021.

\bibitem[Zhao et~al.(2023{\natexlab{b}})Zhao, Bai, Zhang, Zhang, Xu, Lin, Timofte, and Van~Gool]{zhao2023cddfuse}
Zixiang Zhao, Haowen Bai, Jiangshe Zhang, Yulun Zhang, Shuang Xu, Zudi Lin, Radu Timofte, and Luc Van~Gool.
\newblock Cddfuse: Correlation-driven dual-branch feature decomposition for multi-modality image fusion.
\newblock In \emph{Proceedings of the IEEE/CVF conference on computer vision and pattern recognition}, pages 5906--5916, 2023{\natexlab{b}}.

\bibitem[Zhao et~al.(2023{\natexlab{c}})Zhao, Bai, Zhu, Zhang, Xu, Zhang, Zhang, Meng, Timofte, and Van~Gool]{zhao2023ddfm}
Zixiang Zhao, Haowen Bai, Yuanzhi Zhu, Jiangshe Zhang, Shuang Xu, Yulun Zhang, Kai Zhang, Deyu Meng, Radu Timofte, and Luc Van~Gool.
\newblock Ddfm: denoising diffusion model for multi-modality image fusion.
\newblock In \emph{Proceedings of the IEEE/CVF International Conference on Computer Vision}, pages 8082--8093, 2023{\natexlab{c}}.

\bibitem[Zhou et~al.(2023)Zhou, Huang, Wang, Song, and Yang]{zhou2023xnet}
Yanfeng Zhou, Jiaxing Huang, Chenlong Wang, Le Song, and Ge Yang.
\newblock Xnet: Wavelet-based low and high frequency fusion networks for fully-and semi-supervised semantic segmentation of biomedical images.
\newblock In \emph{Proceedings of the IEEE/CVF International Conference on Computer Vision}, pages 21085--21096, 2023.

\end{thebibliography}
